\pdfoutput=1

\documentclass[11pt]{article}

\usepackage[final]{acl}

\usepackage{times}
\usepackage{latexsym}
\usepackage{float}

\usepackage[T1]{fontenc}

\usepackage[utf8]{inputenc}

\usepackage{microtype}

\usepackage{inconsolata}

\usepackage{graphicx}
\usepackage{tabularray}

\usepackage{times}
\usepackage{latexsym}

\usepackage[normalem]{ulem}
\useunder{\uline}{\ul}{}
\usepackage{longtable}
\usepackage{multirow}

\PassOptionsToPackage{dvipsnames, table}{xcolor}
\usepackage{colortbl}
\usepackage{arydshln}
\usepackage{booktabs}
\usepackage{makecell}
\usepackage{fancyvrb}
\usepackage{tcolorbox}
\tcbuselibrary{skins} 
\usepackage{amsmath}
\usepackage{placeins}

\newtcolorbox{dialogbox}{
    enhanced,
    boxrule=1pt, 
    colback=black!10, 
    colframe=black, 
    left=1pt, 
    right=1pt, 
    top=3pt, 
    bottom=3pt, 
}

\newcommand{\wrong}[1]{\underline{\textit{#1}}}
\title{CAST: Corpus-Aware Self-similarity Enhanced Topic modelling}

\author{
 \textbf{Yanan Ma$^{1}$}\thanks{\quad Corresponding contact email addresses: \{yanan.ma, chenhan.yuan, sabine.vanderveer, lamiece.hassan, chenghua.lin, gnenadic\}@manchester.ac.uk, chenghao.xiao@durham.ac.uk. Our code is available at: \url{https://github.com/yananma1029/CAST}.
 }\quad
 \textbf{Chenghao Xiao$^2$}\quad
 \textbf{Chenhan Yuan$^1$}\quad
 \textbf{Sabine N van der Veer$^1$}\quad \\
 \textbf{Lamiece Hassan$^1$}\quad
  \textbf{Chenghua Lin$^1$}\quad
 \textbf{Goran Nenadic$^1$}\\
 $^1$The University of Manchester, $^2$Durham University
}

\begin{document}
\maketitle
\begin{abstract}
Topic modelling is a pivotal unsupervised machine learning technique for extracting valuable insights from large document collections. Existing neural topic modelling methods often encode contextual information of documents, while ignoring contextual details of candidate centroid words, leading to the inaccurate selection of topic words due to the \textit{contextualization gap}. In parallel, it is found that functional words are frequently selected over topical words. To address these limitations, we introduce \textbf{CAST}: \textbf{C}orpus-\textbf{A}ware \textbf{S}elf-similarity Enhanced \textbf{T}opic modelling, a novel topic modelling method that builds upon candidate topic word embeddings contextualized on the dataset, and a novel self-similarity-based method to filter out less meaningful tokens. Inspired by findings in contrastive learning that self-similarities of functional token embeddings in different contexts are much lower than topical tokens, we find self-similarity to be an effective metric to prevent functional words from acting as candidate topic words. Our approach significantly enhances the coherence and diversity of generated topics, as well as the topic model's ability to handle noisy data. Experiments on news benchmark datasets and one Twitter dataset demonstrate the method's superiority in generating coherent, diverse topics, and handling noisy data, outperforming strong baselines.
\end{abstract}
\section{Introduction}

Topic modelling is a powerful tool for extracting valuable insights and detecting topics of interest within a large collection of documents, where the conventional statistical-based models~\citep{lin2012feature,barawi2017automatically,egger2022topic} were typically developed based on the dominant Latent Dirichlet Allocation (LDA) model \citep{blei2003latent}. However, LDA relies on the bag-of-words (BOW) model which has inherent limitations such as ignoring the semantic meaning of words, word order, and contextual information.

\begin{figure*}[ht]
  \centering
  \includegraphics[width=0.8\textwidth]
  {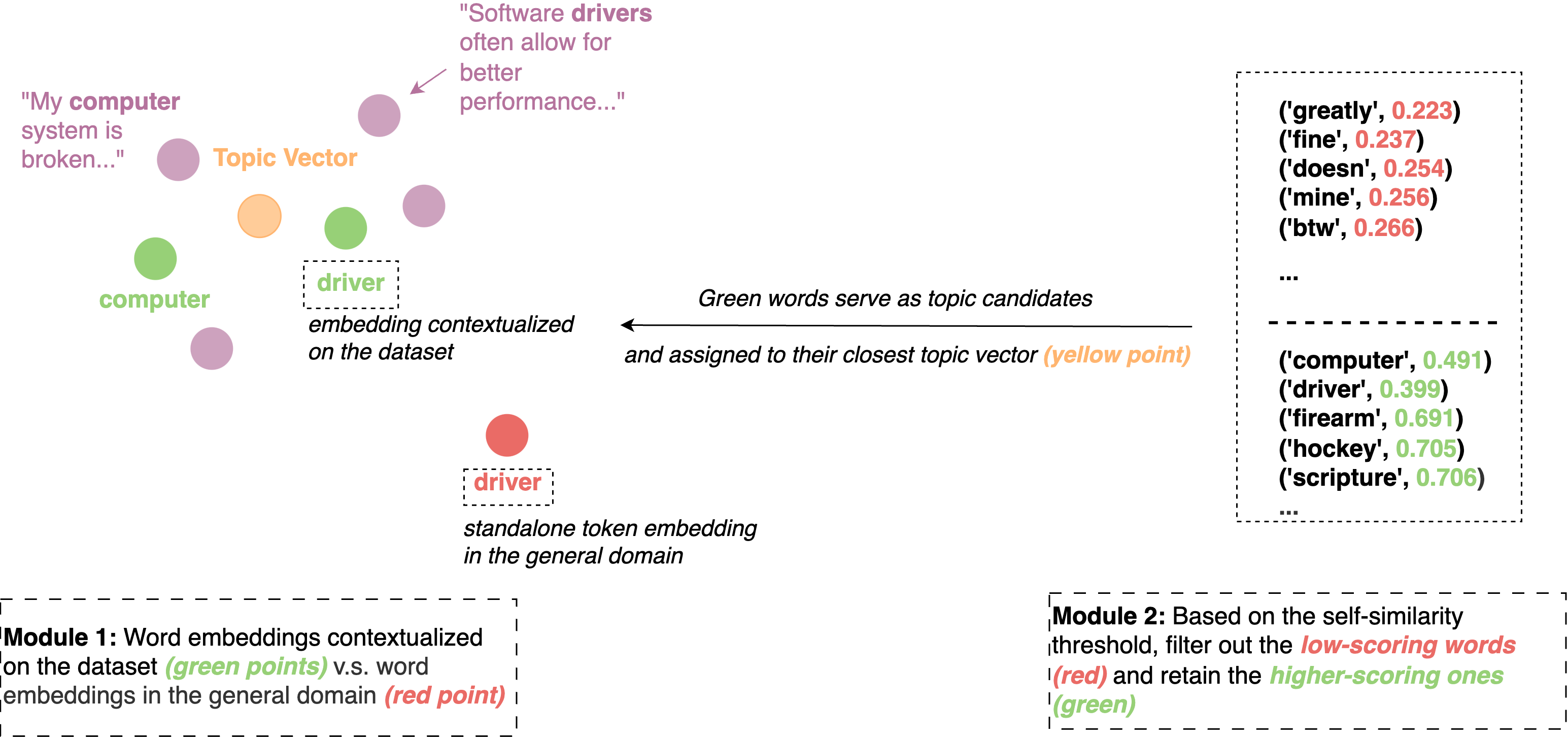}
  \caption{Two modules to identify meaningful candidate topic words. \textbf{Module 1}: word embeddings contextualized on the dataset. \textbf{Module 2}: self-similarity scores to filter out functional words. Purple points represent documents, with semantically similar ones clustered together. Words with higher self-similarity scores (green) are selected over those with lower scores and assigned to their closest cluster centroid (topic vector: yellow point) as topic words (green points), rather than relying on general-domain topic words (red points).}
  \label{fig:ourmodel}
\end{figure*}

To address these issues, recent studies have incorporated embedding techniques, such as word2vec \citep{mikolov2013efficient, mikolov2013distributed}, BERT \citep{devlin2018bert} and Sentence-BERT \citep{reimers2019sentence}, into topic modelling. For instance, \citeauthor{sia2020tired} (\citeyear{sia2020tired}) integrates clustering algorithms with various embedding models to identify topics, demonstrating that this method achieves comparable performance to traditional LDA while reducing complexity and runtime. Similarly, Top2Vec \citep{angelov2020top2vec} and BERTopic \citep{grootendorst2022bertopic} use embedding models to encode and cluster document embeddings based on their semantic similarities, representing each cluster as a topic. These models have shown significant advancements over LDA in previous studies in terms of generating coherent and diverse topics that capture semantic relationships and context in the documents \citep{yuan-eldardiry-2021-unsupervised,angelov2020top2vec, grootendorst2022bertopic, egger2022topic}.

However, a common limitation of the aforementioned models is their failure to fully consider contextual information within documents. Although Top2Vec utilizes contextual embedding models to generate document embeddings, when encoding candidate topic words, it encodes them as standalone words, disregarding surrounding context. This has prevented taking into account the contextualized meanings of these words in the corpus of diverse topics. For instance, the embeddings for \textit{``bank''} in the context of a financial institution and a riverbank would be identical, depending solely on the embedding model they use. This can lead to inaccurate topic word assignments, and failure to properly represent the topics, especially when documents have a domain shift compared to the training corpus of the embedding model. Similarly, BERTopic derives its topic representations from a BOW approach which fails to leverage the advancements of embedding techniques. This method may have limitations such as neglecting the contextual information captured by the language model, which can result in missing crucial semantic nuances and word relationships within topics.

Another area where current topic modelling methods fall short lies in building meaningful topic words. Topic models like LDA and BERTopic may inadvertently prioritize common terms and sometimes select functional words, such as stopwords, over topical words, despite extensive pre-processing. Standard stopword lists, like those from \texttt{NLTK}, may not fully capture all irrelevant terms for a specific corpus and often require prior knowledge to build a customized stopword list tailored to the dataset. 

To address the above limitations, we introduce \textbf{CAST}: \textbf{C}orpus-\textbf{A}ware \textbf{S}elf-similarity Enhanced \textbf{T}opic modelling, a novel topic model that extends Top2Vec by introducing two novel modules: 1) word embeddings contextualized on the dataset and 2) self-similarity scores to filter out functional words and construct meaningful candidate topic words, as shown in Figure \ref{fig:ourmodel}.

More specifically, we first leverage embedding models to joint encode word and document embeddings. Candidate topic words are encoded based on their usage within the documents, instead of statically encoded in the general domain as standalone tokens (the red point). Second, we reduce the dimensionality of the document embeddings and cluster the reduced embeddings to create semantically similar topic clusters (purple points). Third, inspired by the findings in contrastive learning that functional tokens have lower self-similarity scores than that of semantic tokens \cite{xiao2023isotropy}, we employ a self-similarity threshold to effectively filter out functional words or less relevant words and keep meaningful candidate topic words. Finally, we assign the candidate words closest to their cluster centroid (yellow point) as the corresponding topic words (green points).

The contributions of this paper are three-fold: 
\begin{itemize}
\item[1)] We propose a novel topic modelling method that takes into account corpus-level contextual information when encoding candidate topic words, pushing closer the candidate topic word embeddings with the contextualized topic vectors.
\item[2)] We introduce the novel use of self-similarity of varied token types in topic modelling, to filter out functional tokens and keep meaningful words as candidate topic words.
\item[3)] We conducted a comprehensive evaluation of CAST using automatic and large language model-based metrics on two benchmark news datasets and one Twitter dataset. CAST shows superior performance in generating coherent and diverse topics, as well as handling noisy data compared with the strong baselines. 
\end{itemize}

\section{Contextualized Topic modelling}
\subsection{Candidate Topic Word Embeddings Contextualized on Dataset}

Existing neural topic modelling methods like Top2Vec encode document embeddings, taking into account their contextualized meanings. However, when selecting candidate topic words, they encode individual words without their contexts. We posit that this results in a \textit{Contextualization gap}, illustrated in Figure \ref{fig:ourmodel} (left), i.e., the candidate topic words are not contextualized in any context.

In contrast, we contextualize candidate word embeddings based on the corpus to be modelled, e.g., the embedding of \textit{``driver''} would be the average word embeddings contextualized in actual documents across the corpus. The resulting embedding would be different from the embedding of \textit{``driver''} encoded in a standalone manner.

In CAST, we first extract the word embeddings from different documents across the corpus and then average them to obtain the final word embedding. In this way, the word embedding can preserve the contextual information in the corpus and capture the nuances and variations in word meanings that arise from different contexts. 

The final word embedding, $E_{\text{final}}$, is defined as the average of the contextualized word embeddings of a word $w$ across all sequences in a corpus. Formally:
\begin{equation}
E_{\text{final}} = \frac{1}{P} \sum_{i=1}^{P} C_{wi}
\end{equation}
where $C_{wi}$ denotes the contextualized word embedding of the word $w$ in the $i$-th sequence. $P$ is the number of sequences that contain $w$.

Given a sentence $S$, the following steps are performed to compute $C_{wi}$:

\begin{itemize}
    \item[1.] Tokenize $S$ into subwords $\{t_1, t_2, \dots, t_n\}$.
    \item[2.] For each word $w$, composed of subwords $\{t_{i_1}, t_{i_2}, \dots, t_{i_k}\}$ (e.g. \texttt{``happi''}, \texttt{``\#\#ness''}), its $C_{wi}$ is calculated as:
    \[
    C_{wi} = \frac{mean\left( \mathbf{e}_{i_1}, \mathbf{e}_{i_2}, \ldots, \mathbf{e}_{i_k} \right)}{\left| mean\left( \mathbf{e}_{i_1}, \mathbf{e}_{i_2}, \ldots, \mathbf{e}_{i_k} \right) \right|}
    \]
\end{itemize}

where $mean$ denotes the element-wise mean of the subword embeddings, and $|\cdot|$ represents the $L2$ norm.

\subsection{Self-similarity Scores}
\label{sec: SS score}

In contrast to LDA and BERTopic which rely on preprocessing for filtering out functional words, we adopt Top2Vec's method, which assigns topic candidates to their closest cluster centroids as topic words. Common words, which are equally distant from all clusters, tend to be far from any centroid and are thus discarded. We extend this approach by adding a module to calculate word self-similarity scores across documents, enabling further automatic filtering out of functional words in a post-processing step. This is inspired by the observation of \citet{xiao2023isotropy} that after contrastive learning, self-similarities of semantic words become higher than a model's pre-trained checkpoint, while self-similarities of functional words become lower. This is referred to as \textit{``entourage effect''} \cite{xiao2023length}, highlighting that functional word embeddings tend to follow the movement of semantic word embeddings from the same sentence in the embedding space, making the self-similarity for the functional word lower across the documents. This pattern grants us the natural convenience of filtering functional words out - as these are the ones that have lower self-similarities. Given that almost all sentence embedding models used in topic modelling are facilitated by contrastive learning, the pattern becomes universally useful in facilitating our framework.

In CAST, we calculate each word' cosine-similarity with itself as its self-similarity score. For a word \(w\), its self-similarity score \(SS_w\) is given by 
\begin{equation}
SS_w = \text{cosine\_similarity}([E_w], [E_w])
\end{equation}
where $E_w = \{ C_{w1}, C_{w2}, \dots, C_{wi} \}$ is a list of contextualized word embeddings of word $w$ across the corpus.

We then sort the self-similarity scores and manually select a threshold to exclude most functional words (those with lower self-similarity scores) while retaining meaningful words as candidate topic words. For example, setting \texttt{threshold = 0.3} could exclude functional words such as \textit{``<pad>''} and \textit{``due''} (see Table \ref{tab:self-similarity scores}). 

\subsection{Document Clustering}
In embedding spaces, document vectors with semantically similar concepts tend to cluster together. However, in high-dimensional embedding spaces, data becomes sparse, and the distance differences between nearest and farthest neighbours diminish, complicating the identification of meaningful clusters \citep{beyer1999nearest}. This phenomenon is referred to as the ``\textit{curse of dimensionality}'' \citep{Bellman+1961}.

To mitigate this issue, we employed Uniform Manifold Approximation and Projection (UMAP) \citep{mcinnes2018umap} to reduce the dimensionality of the embeddings and applied HDBSCAN \citep{campello2013density} to cluster the document embeddings in the reduced dimensions. 

\subsection{Topic Representations}
We use the document vectors in the original high-dimensional embedding space to represent the documents, as they preserve all the information. Assuming documents from the same cluster represent the same topic, we calculate the centroid vector of the document embeddings in each cluster to serve as the topic vector. Candidate topic words are then assigned to their nearest topic vectors based on cosine similarities.

\section{Experiment}
\subsection{Dataset and Preprocessing}

In our paper, we evaluated CAST on three benchmark datasets, namely, 20NewsGroups, BBC News datasets, and Elon Musk's tweets. The data statistics can be found in Table~\ref{data-statistics}. 

We retrieved the preprocessed versions of 20NewsGroups and BBC News datasets from the OCTIS library \footnote{\url{https://github.com/MIND-Lab/OCTIS?tab=readme-ov-file}} \citep{terragni2021octis} with punctuation and stopwords removed and words lemmatized in the main experiments. 

To test the ability of topic models to identify the nuance between topics in a short-text and social media setting, we retrieved the publicly available Elon Musk's tweets dataset\footnote{\url{https://www.kaggle.com/datasets/yasirabdaali/elon-musk-tweets-dataset-17k}} and included 13,998 original tweets after excluding tweets that only contain links. We removed all the mentions (e.g. @elonmusk), hashtags, and unknown characters in the dataset and further preprocessed it by removing stop-words and lemmatization using OCTIS.

\subsection{Baselines}
We evaluated CAST against the traditional LDA and three recently developed models:\\
\textbf{LDA} is a widely used topic modelling technique that serves as a baseline for comparison with other models. It represents each document as a mixture of topics and each topic as a probabilistic distribution of words.  \\
\textbf{Top2Vec} employs word embeddings and clustering methods by encoding documents and words and clustering document embeddings into semantically related dense areas. Then it assigns words whose embeddings are close to the centroid of each cluster as topic words. \\
\textbf{BERTopic} employs a hybrid approach that combines embedding models and a class-based tf-idf method for topic word assignment. Similar to Top2Vec, it encodes and clusters the document embeddings. For each cluster, however, it diverges by using tf-idf to identify topic words from documents within the same clusters. \\
\textbf{TopClus} employs two modules namely the attention-based document embedding learning module and a latent space generative module to jointly learn topics and document embeddings in a shared latent space. The learned topic-word and document-topic distributions represent topics and their associations with documents, while the attention mechanism captures the varying importance of words for topic discovery \citep{meng2022topic}.

In our experiment, we loaded the default LDA from the OCTIS library and used the default parameters of TopClus. We employed two sentence-transformers, \texttt{all-MiniLM-L6-v2} and \texttt{all-mpnet-base-v2}, to run Top2Vec, BERTopic, and CAST. We reduced the embedding dimensionality to five and adjusted the \texttt{min\_cluster\_size} and the \texttt{\#Topics} based on the dataset size (see Table \ref{data-statistics}), with larger datasets having a larger \texttt{min\_cluster\_size} value and more topics to detect. We kept and fixed the other parameters for UMAP and HDBSCAN across all models for a fair comparison. For each model, all the results were averaged across five independent runs across three different numbers of topics in the datasets (15 runs per model in total) by using one NVIDIA A100 GPU. See Appendix \ref{appen: implementation} for a more detailed implementation description.

\noindent
\begin{table*}[ht]
\centering
\scalebox{0.7}{
\begin{tabular}{ccccccccccccc}
\toprule
\multirow{2}{*}{\textbf{Models}} &\multicolumn{4}{c}{\textbf{20NewsGroups}} &\multicolumn{4}{c}{\textbf{BBC News}} &\multicolumn{4}{c}{\textbf{Elon Musk}} \\
\cmidrule(lr){2-5} \cmidrule(lr){6-9} \cmidrule(lr){10-13}
&TC & TD & LLM-TC & LLM-TD & TC & TD & LLM-TC & LLM-TD & TC & TD & LLM-TC & LLM-TD \\
\midrule
LDA & $0.06$ & $0.74$ & $2.88$ & $2.6$  & $0.02$ & $0.68$ & $2.79$ & $2$ &  $-0.1$ & $0.79$ & $2.67$ & $2.8$\\
TopClus & $-0.01$ & \cellcolor{green!80}$1$ & $3.13$ & \cellcolor{green!30}$3.6$ & $-0.32$ & \cellcolor{green!80}$1$ & $2.69$ & \cellcolor{green!80}$3.4$ & $-0.39$ & \cellcolor{green!80}$0.99$ & $2.81$ & \cellcolor{green!80}$3.4$  \\
BRTo-\textit{MiniLM} & $0.07$ & $0.83$ & $3.2$ & $2.6$ & \cellcolor{green!30}$0.13$ & $0.89$ & $3.68$ & \cellcolor{green!30}$3.2$ & $-0.12$ & \cellcolor{green!30}$0.96$ & $2.85$ & \cellcolor{green!30}$3.2$ \\
BRTo-\textit{MPNET} & $0.11$ & $0.81$ & $3.2$ & $3$ & \cellcolor{green!80}$0.14$ & $0.89$ & $3.71$ & \cellcolor{green!30}$3.2$ & $-0.1$ & $0.93$ & $2.85$ & \cellcolor{green!80}$3.4$\\
T2V-\textit{MiniLM} & $0.1$ & $0.85$ & $3.59$ & $2.8$ & $-0.01$ & $0.85$ & $3.77$ & \cellcolor{green!80}$3.4$ & $-0.13$ & $0.72$ & $3.1$ & $2.8$\\
T2V-\textit{MPNET} & $0.07$ & $0.78$ & $3.54$ & $2.6$ & $-0.05$ & $0.83$ & $3.75$ & $3$ & $-0.12$ & $0.68$ & $3.05$ & $2.4$\\
CAST-\textit{MiniLM} & $0.12$ & $0.87$ & \cellcolor{green!30}$3.77$ & \cellcolor{green!80}$3.8$ & \cellcolor{green!30}$0.13$ & $0.86$ & \cellcolor{green!80}$3.89$ & $3$ & $-0.06$ & $0.62$ & \cellcolor{green!30}$3.32$ & $3$\\
CAST-\textit{MPNET} & \cellcolor{green!80}$0.18$ & $0.9$ & \cellcolor{green!80}$3.88$ & \cellcolor{green!80}$3.8$ & $0.12$ & \cellcolor{green!30}$0.94$ & \cellcolor{green!30}$3.78$ & $3$ & \cellcolor{green!80}$-0.03$ & $0.73$ & \cellcolor{green!80}$3.39$ & $2.8$ \\

\midrule
CAST-\textit{MiniLM} $^*$ & $0.1$ & $0.82$ & $3.61$ & \cellcolor{green!30}$3.6$ & $0.12$ & $0.82$ & \cellcolor{green!30}$3.78$ & $2.4$ & $-0.08$ & $0.62$ & $3.05$ & $2.4$\\
CAST-\textit{MPNET} $^*$ & \cellcolor{green!30}$0.17$ & \cellcolor{green!30}$0.93$ & $3.72$ & \cellcolor{green!30}$3.6$ & $0.11$ & $0.93$ & $3.77$ & $3$ & \cellcolor{green!30}$-0.05$ & $0.75$ & $3.2$ & \cellcolor{green!80}$3.4$\\ 

\bottomrule
\end{tabular}}
\caption{We evaluated all models using traditional topic coherence (TC) and topic diversity (TD), alongside large language model-based metrics: LLM-TC and LLM-TD. For each model, all the results were averaged across five independent runs across three different numbers of topics in the datasets (15 runs per model in total). The best two values under each metric were highlighted in green with darker colors indicating greater value. $^*$ means the vanilla model that does not use the self-similarity module to filter words.}
\label{tab:avg_score}
\vspace{-5mm}
\end{table*}
\subsection{Evaluation Metrics}
\textbf{Quantitative evaluation.} Topic coherence (TC) and topic diversity (TD) are two widely adopted quantitative metrics for evaluating the performance of topic models. An effective topic model should generate coherent and semantically meaningful words to represent each topic, while also ensuring diversity across the different topics. In our paper, we used both automatic metrics and large language models (LLMs) based methods.

(1) \textit{Automatic metrics}: We used Normalized Point-wise Mutual Information (NPMI) \citep{bouma2009normalized} to assess the TC as it has shown the ability to emulate human judgment \citep{lau-etal-2014-machine}. This method relies on the word co-occurrence in the corpus and ranges from -1 to 1, with 1 indicating a strong positive coherence and -1 a strong negative coherence.  We reported TD using the percentage of unique words in the top words of all topics, as defined in \citep{dieng2020topic}. This metric ranges from 0 to 1, with higher scores representing more diverse topics. 

(2)\textit{ LLM-based metrics}: Human evaluation is widely regarded as the gold standard for assessing topic models \citep{grootendorst2022bertopic, rahimi-etal-2024-contextualized}. However, this process is resource-intensive. In our study, we used GPT-4 to mimic the human evaluation. Following \citeauthor{newman2010automatic} (\citeyear{newman2010automatic}), we developed prompts for Topic Coherence (TC) and Topic Diversity (TD) assessment using 4-point rating scales (Figure \ref{fig:prompt}). To validate this approach, we conducted a human evaluation on a subset of the 20NewsGroups results. The Pearson correlation coefficients between GPT-4 and human evaluations were 0.869 for TC and 0.732 for TD, supporting the efficacy of GPT-4 for automatic evaluation. Detailed methodology is provided in Appendix \ref{appen: GPT-4 validation}.

\noindent\textbf{Qualitative evaluation.} We further examined the ability of topic models to generate meaningful words over noisy words as a metric to evaluate the models. To achieve this, we randomly selected five labels and manually matched the most relevant topics generated by all models. We then manually reviewed the topic words, underlining and italicizing those irrelevant to the topic, which we identified as noisy words (see Table \ref{tab:topicwords}). We only conducted the qualitative evaluation on 20NewsGroups and BBC News datasets, as they are human-labelled datasets with topic labels (e.g. sports, politics).

\noindent
\setlength{\tabcolsep}{3pt}
\begin{table*}[ht]
\centering
\vspace*{-1em}
\resizebox{\textwidth}{!}{
\begin{tabular}{ccccccccccc}
\toprule
\multirow{3}{*}{Models} &
\multicolumn{5}{c}{\textbf{20NewsGroups}} & \multicolumn{5}{c}{\textbf{BBC News}} \\
\cmidrule(lr){2-6} \cmidrule(lr){7-11}
& Topic 1 & Topic 2 & Topic 3 & Topic 4 & Topic 5 & Topic 1 & Topic 2 & Topic 3 & Topic 4 & Topic 5\\
& (Computer) & (Auto) & (Sport) & (Space) & (Mideast) & (Tech) & (Business) & (Sport) & (Entertainment) & (Politics) \\

\midrule
\multirow{5}{*}{\makecell{LDA}} 
& image & car & game & space & arab & phone & growth & game & music & labour \\
& color & \wrong{good} & team & \wrong{year} & isracli & service & economy & player & network & election \\
& bit & owner & play & science & wall & video & sale & \wrong{company} & film & government \\
& card & buy & player & research & panel & user & rise & team & work & party \\
& file & gun & \wrong{year} & launch & hot & mobile & price & club & \wrong{good} & \wrong{company} \\

\midrule
\multirow{5}{*}{\makecell{TopClus}} 
& dat & {} & {} & flight & german & desktop & restructuring & finalist & cult & pun \\
& del & {} & {} & ground & criminal & blog & recession & squad & horror & tel \\
& graphic & $-$ & $-$ & water & terrorist & server & profit & bronze & theatre & police \\ 
& \wrong{black} & {} & {} & \wrong{wall} & armed & portal & warning & sprinter & fame & constituency \\
& image & {} & {} & oil & catholic & browser & currency & champion & drama & democratic \\

\midrule
\multirow{5}{*}{\makecell{BERTopic-\textit{MiniLM}}} 
& monitor & car & game & \wrong{car} & {} & mobile & price & win & film & government \\
& driver & drive & team & \wrong{drive} & {} & phone & rise & play & award & election\\
& color & \wrong{space} & play & space & $-$ & service & oil & game & \wrong{good} & party \\
& card & bike & player & \wrong{bike} & {} & technology & sale & player & show & labour \\
& vga & \wrong{launch} & year & launch & {} & user & market & match & star & tory \\

\midrule
\multirow{5}{*}{\makecell{BERTopic-\textit{MPNET}}} 
& monitor & car & game & {} & armenian & game & company & win & film & government \\
& card & bike & team & {} & people & phone & firm & play & \wrong{good} & election \\
& video & ride & play & $-$ & turkish & mobile & share & game & award & party \\
& driver & engine & player & {} & government & technology & sale & player & actor & labour \\
& vga & \wrong{good} & win & {} & israeli & user & profit & match & star & plan \\

\midrule
\multirow{5}{*}{\makecell{Top2Vec-\textit{MiniLM}}} 
& vga & vehicle & baseball & {} & isracli & devices & irms & premiership & oscar & tory \\
& monitor & car & league & {} & genocide & mobiles & takeover & mourinho & cast & blair \\
& hardware & engine & hockey & $-$ & armenian & phones & shareholders & hodgson & oscars & ukip \\
& motherboard & truck & playoff & {} & peace & digital & stock & liverpool & cinema & mps \\
& screen & dealer & pitch & {} & massacre & mobile & investors & gerrard & actor & minister \\

\midrule
\multirow{5}{*}{\makecell{Top2Vec-\textit{MPNET}}} 
& vga & vehicle & baseball & orbit & genocide & devices & shareholders & nadal & mtv & blair \\
& monitor & car & hockey & space & israeli & mobiles & companies & federer & songs & tory \\
& display & engine & pitch & shuttle & arab & phones & firms & tennis & singer & labour \\
& screen & auto & league & engineering & terrorist & technology & investors & iaaf & album & tories \\
& resolution & motorcycle & sport & implement & conflict & telecoms & finance & olympic & concert & ukip \\

\midrule
\multirow{5}{*}{\makecell{Ours-\textit{MiniLM}}}
& driver & car & game & mission & israeli & mobiles & alleged & wales & singer & tories \\
& card & vehicle & team & orbit & arab & mobile & allegations & rugby & music & tory \\
& chip & truck & baseball & space & peace & phones & sec & squad & band & mps \\
& hardware & motor & player & project & jewish & phone & fraud & england & rock & ministers \\
& vga & auto & score & launch & oppose & telecoms & executives & football & songs & parliamentary \\

\midrule
\multirow{5}{*}{\makecell{Ours-\textit{MPNET}}}
& monitor & car & game & launch & occupy & gadgets & takeover & mourinho & oscars & imf \\
& vga & rear & score & space & territory & gadget & boerse & liverpool & nominations & economist \\
& video & vehicle & team & mission & soldier & portable & lse & wenger & festival & turkey \\
& screen & brake & play & solar & peace & mobiles & stake & gerrard & documentary & poverty \\
& display & engine & player & flight & Israeli & handheld & mci & chelsea & dvd & reforms \\

\bottomrule
\end{tabular}
}
\caption{Qualitative evaluation of topic discovery. We randomly selected five labels from 20NewsGroups and BBC News and manually matched the most relevant topics generated by all models. Words not strictly belonging to the corresponding topic are italicized and underlined as noisy words.}
\label{tab:topicwords}
\vspace*{-1.5em}
\end{table*}

\section{Results}

The main results are shown in Table \ref{tab:avg_score} and Table \ref{tab:topicwords}. 

\noindent\textbf{Quantitative results.} From Table \ref{tab:avg_score}, we can see that CAST models generally generate more coherence and diverse topics than baseline models. 

In terms of TC, CAST models exhibit consistently high TC scores on both traditional NPMI and LLM-based coherence metrics across all datasets. Among the two variants, CAST-MPNET achieves the highest NPMI and LLM-TC scores on the 20NewsGroups and Elon Musk datasets. CAST-MiniLM demonstrates consistent and competitive coherence on all datasets, with the best LLM-TC score on BBC News, and the second-best on 20NewsGroups.

In terms of TD, TopClus stands out in TD across all datasets. CAST shows competitive performance compared to TopClus, with the best LLM-TD on 20NewsGroups, indicating a balance between topic coherence and diversity.

The vanilla CAST models (marked with *), which ablate the self-similarity module, still outperform baseline models in most cases, underscoring the efficacy of the contextualized embedding module to improve the model performance. However, compared with the full CAST models, the vanilla models exhibit marginally lower coherence scores while maintaining robust diversity metrics. This suggests that the self-similarity module, by filtering out functional words, could enhance coherence at a slight cost to topic word diversity. We further explain this in Section \ref{sec: ablation}.

\noindent\textbf{Qualitative results.} Table \ref{tab:topicwords} further analyzes models' ability to generate meaningful topic words over irrelevant words. In general, Top2Vec and CAST are consistent in generating coherent and contextually relevant topic words. However, Top2Vec fails to identify Topic 4 (Space) on 20NewsGroups. Furthermore, Table \ref{tab:detailed_twords} reveals that Top2Vec-MPNET occasionally generates less meaningful topic words, such as \textit{``inform''}, \textit{``topic''}, and \textit{``implement''} in the first two topics, and the second topic is too ambiguous, making it hard to distinguish between \texttt{medicine} and \texttt{business}. This is potentially due to Top2Vec's approach of merging the smallest cluster with the most similar ones to achieve the desired number of topics. While this method can effectively merge similar topics, it could also lead to pollution to larger clusters and potentially ignore certain topics. In contrast, CAST selects the top-$n$ clusters according to their sizes to determine the desired number of topics. This approach ensures that significant topics are preserved and minimizes the risk of smaller clusters distorting the embeddings of larger, meaningful clusters.

In terms of LDA and BERTopic, they frequently generate noisy and semantically irrelevant words despite their limited meaningfulness. This is potentially because these two models do not leverage the semantic richness offered by word embeddings to identify topic words. TopClus and BERTopic models exhibit significant gaps in topic coverage. Specifically, BERTopic struggles to distinguish between Topics 2 and 4 in the 20NewsGroups dataset, leading to a conflation of topic words. This indicates a limitation in the model's ability to maintain distinct topic boundaries. TopClus tends to cluster meaningless adjectives and verbs together as a topic. For example, \textit{``good'', ``long'', ``big'', ``small'', ``large''} in Table \ref{tab:detailed_twords}.

\noindent
\begin{table*}[ht]\centering
\scalebox{0.8}{
\begin{tabular}{cccccl}
\toprule
\multicolumn{1}{c}{\textbf{Models ↓  Metrics→}} &\multicolumn{1}{c}{\textbf{NPMI}} &\multicolumn{1}{c}{\textbf{TD}} &\multicolumn{1}{c}{\textbf{LLM-TC}} &\multicolumn{1}{c}{\textbf{LLM-TD}} & \multicolumn{1}{c}{\textbf{Topic Words Examples}} \\
\midrule
LDA	& $-0.016$ & $0.623$ &	$1.6$	& $1.333$ &	[`>', `ax', `\#', `@', `('] \\
TopClus &	$-0.237$	 & $\textbf{1}$ &	$2.4$ &	$\textbf{4}$	& [`travels', `consists', `provides', `lacks', `wears']\\
BERTopic-\textit{MPNET} & $\underline{0.028}$ &	$0.323$	& $1.63$	& $\underline{1.667}$	& [`the', `to', `and', `is', `of']\\
Top2Vec-\textit{MPNET}	& $-0.152$	& $0.977$	& $\underline{3.53}$	& $\textbf{4}$	& [`vehicles', `vehicle', `motorcycles', `honda', `mustang'] \\
 CAST-\textit{MPNET} & $\textbf{0.037}$ & $\underline{0.99}$	& $\textbf{3.7}$	& $\textbf{4}$	& [`cars', `autos', `car', `vehicles', `toyota']\\
\bottomrule
\end{tabular}}
\caption{We tested the models' performance in dealing with noisy data on the non-preprocessed 20 Newsgroups dataset fetched from \texttt{sklearn}, with the number of topics set to 10. Results were averaged across five independent runs for each model with the best value in each metric highlighted in bold and the second best underlined. We manually chose one typical topic with five topic words to better show the models' ability to deal with noisy data.}
\label{tab:noisy data}
\vspace{-3mm}
\end{table*}

\begin{figure}[ht]
  \includegraphics[width=\columnwidth]{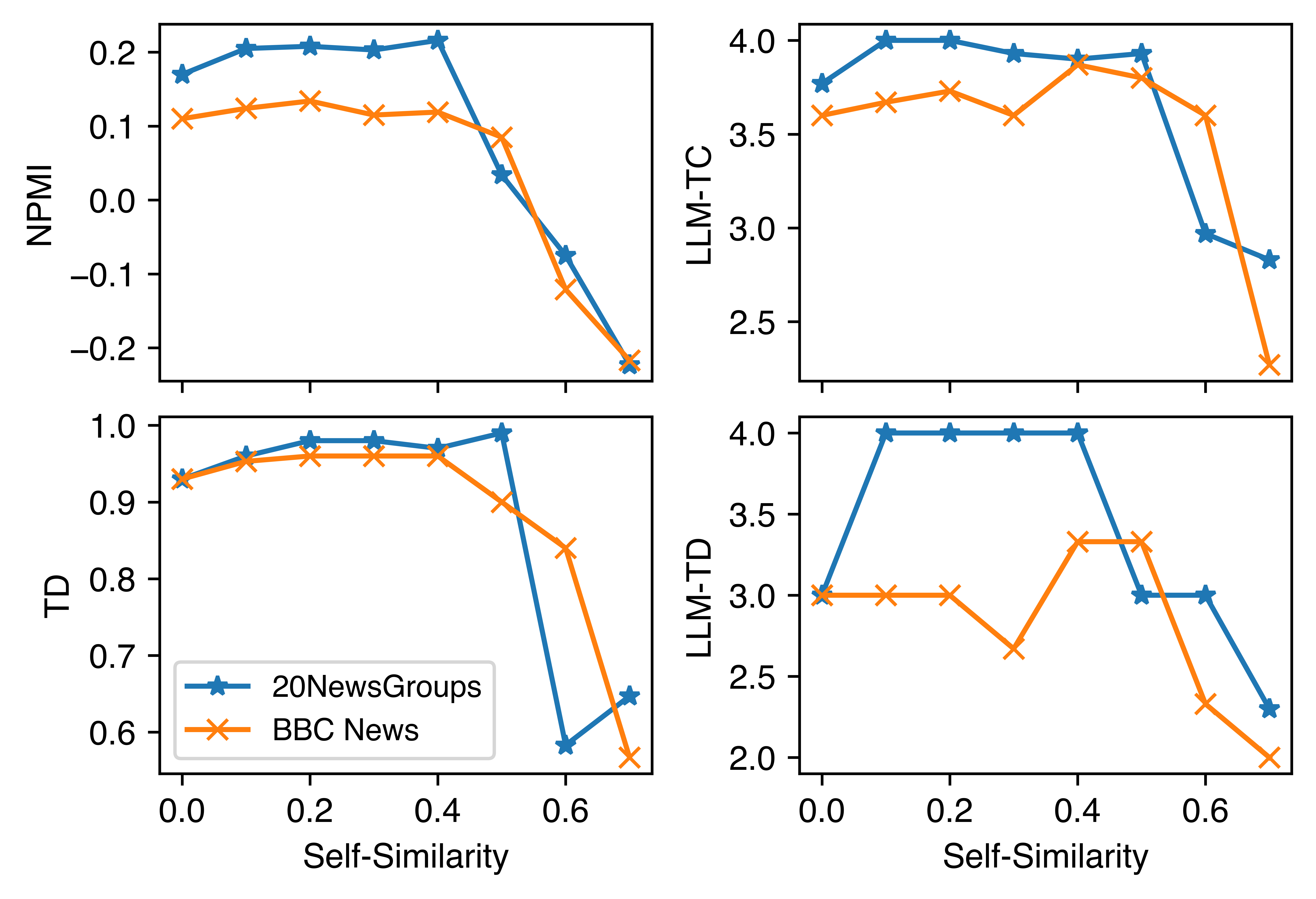}
  \caption{Ablation analysis of varying self-similarity thresholds on CAST-MPNET for the 20NewsGroups (\texttt{nr\_topics = 10}) and BBC News (\texttt{nr\_topics = 5}) datasets. All results were averaged across five independent runs for each threshold value. The analysis was constrained to a maximum threshold of 0.7 because the model could not identify sufficient topic words after this threshold.}
  \label{fig:ablation}
\vspace{-5mm}
\end{figure}

\subsection{Dealing with Noisy Data}
\label{sec: noisy data}
In real-world settings, data are full of noise. Therefore, we evaluated the performance of the models in handling noisy data in this section. Table \ref{tab:noisy data} illustrates the results of CAST-MPNET and its corresponding baseline models tested on the non-preprocessed 20 Newsgroups dataset\footnote{\url{https://scikit-learn.org/0.15/datasets/twenty_newsgroups.html}} as an example.

The results demonstrate that LDA and BERTopic struggle to effectively handle noisy data due to their limitations for topic word assignment. Similarly, TopClus, despite its ability to partially mitigate this issue, tends to cluster functional words, such as verbs and adjectives, as topic words, which may not accurately capture the underlying topics. In contrast, Top2Vec and CAST exhibit robustness in extracting meaningful topic words from noisy data. Both methods do not need preprocessing to identify topic words, as common words—equidistant from all clusters—are typically far from any centroid and are thus discarded (see Section \ref{sec: SS score}). However, some less common function words such as \textit{``inform''} and \textit{``implement''} may still be wrongly selected by Top2Vec (see Table \ref{tab:detailed_twords}). CAST extends Top2Vec with an automatic post-processing self-similarity module, which further removes less meaningful function words. For example, the self-similarity scores for \textit{``inform''} and \textit{``implement''} are 0.218 and 0.26 respectively, and thus are excluded when setting the self-similarity threshold to be 0.4. We further explain the impact of varying thresholds on CAST's performance in Section \ref{sec: ablation}. 

On the other hand, while BERTopic performed second-best on NPMI, its LLM-TC score is quite low, as evident from the topic word examples. This shows that meaningless topics comprising several stopwords may sometimes receive higher NPMI scores due to the common co-occurrence of stopwords like \textit{``the'', ``to'', ``and'',  ``is''} within documents. This observation aligns with previous criticisms that word co-occurrence-based metrics, such as NPMI, are incompatible with recently developed neural topic models \citep{hoyle2021automated,doogan2021topic}. Therefore, it is crucial to use suitable coherence metrics to evaluate recently developed neural topic models, where a potential way is to use LLMs to evaluate the models, as shown in the paper.

\noindent
\begin{table}[!t]
\centering
\resizebox{\columnwidth}{!}{
\begin{tabular}{llll}
\toprule
\textbf{Top 10 SS Score} & \textbf{Threshold=0.5} & \textbf{Threshold=0.4} & \textbf{Threshold=0.3} \\
\midrule
armenian:$0.833$ & student:$0.499$ & capability:$0.399$ & <pad>:$0.299$ \\
genocide:$0.781$ & heart:$0.499$ & driver:$0.399$ & great:$0.299$ \\
turkish:$0.772$ & score:$0.499$ & operation:$0.399$ & stay:$0.298$ \\
homosexuality:$0.764$ & split:$0.499$ & medium:$0.399$ & setting:$0.298$ \\
atheism:$0.755$ & motor:$0.499$ & practice:$0.399$ & good:$0.297$ \\
arab:$0.753$ & legitimate:$0.499$ & confirm:$0.399$ & highly:$0.297$ \\
encryption:$0.738$ & traditional:$0.499$ & exit:$0.399$ & possibly:$0.296$ \\
massacre:$0.737$ & fact:$0.498$ & call:$0.398$ & extremely:$0.295$ \\
homosexual:$0.735$ & sequence:$0.498$ & error:$0.398$ & due:$0.294$ \\
israeli:$0.732$ & software:$0.498$ & default:$0.398$ & advance:$0.293$ \\
\bottomrule
\end{tabular}
}
\caption{Example of words and their self-similarity (SS) scores from the 20NewsGroups dataset. We present the top 10 words with the highest SS scores, along with the top 10 words whose SS scores fall below three thresholds: 0.5, 0.4, and 0.3.}
\label{tab:self-similarity scores}
\vspace*{-1em}
\end{table}

\section{Ablation Analysis}
\label{sec: ablation}

In this section, we ablate the self-similarity thresholds to evaluate how varying thresholds affect TC and TD. We tested CAST-MPNET on the 20NewsGroups and BBC News datasets.

As shown in Figure \ref{fig:ablation}, increasing the self-similarity threshold initially enhances the model's performance, but experiences a marked decline after a higher threshold. As tested on the 20NewsGroups and BBC News datasets, TC and LLM-TC remain competitive and peak around a threshold of 0.4 in most cases, while TD and LLM-TD remain stable with minor fluctuations up to this point. However, beyond a self-similarity score of 0.5, the coherence and diversity metrics experience a significant decline. This pattern indicates that an optimal range of self-similarity scores can increase topic coherence and diversity, but excessively high thresholds may filter out meaningful words, thereby reducing the performance. 

In practical applications, determining the appropriate self-similarity threshold is crucial and will vary depending on the corpus and the embedding model used. In our experiment, we sorted and displayed the self-similarity scores of words across the corpus, and then manually selected a threshold that can effectively exclude most functional words while retaining the most meaningful ones. As shown in Table \ref{tab:self-similarity scores}, a threshold around 0.4 for 20NewsGroup can balance this, thereby, achieving better model performance as verified in Figure \ref{fig:ablation}. 

Additionally, when setting self-similarity to 0, ablating the use of the self-similarity module, CAST still outperforms other models under the same conditions in most cases (see the vanilla models in Table \ref{tab:avg_score}). This demonstrates the efficacy of the contextual embedding method employed by CAST in enhancing the identification of coherent and diverse topics.

\section{Related Work}
\textbf{Neural Topic models.} 
Recent neural topic models such as Top2Vec and BERTopic combine embedding techniques with clustering algorithms. They have demonstrated superior performance in identifying coherent and diverse topics \citep{angelov2020top2vec, grootendorst2022bertopic, egger2022topic}, and have been widely used in various domains \citep{ghasiya2021investigating, jeon2023exploring, wang2023identifying}. 

Our approach uses similar methods as Top2Vec and BERTopic for dimensionality reduction and embedding clustering. However, it differs in attaining the candidate topic word embeddings by averaging the embeddings of the word in different contexts across the corpus, to capture contextual information of the corpus. In contrast, Top2Vec typically extracts static word embeddings from pre-trained embeddings. BERTopic, instead of building word embeddings and finding candidate words whose embeddings are close to the topic embeddings, identifies topic words through a clustering-based tf-idf approach. Furthermore, our method introduces self-similarity scores as a novel mechanism for filtering candidate topic words. Typically, functional words exhibit lower self-similarity scores in their embeddings compared to semantically meaningful words. By establishing an appropriate threshold, we can effectively exclude most functional words while retaining semantically significant terms. This could enhance topic coherence and improve interpretability for end-users.

\textbf{LLM-based topic models.} Recently LLM-based topic models \citep{mu2024large, wang2023prompting}, typically relying on prompt-based strategies to guide LLMs to identify latent topics, have shown strong potential for generating coherent topics. Our approach is complementary to the LLM-based methods by offering a lightweight and cost-effective alternative. By leveraging pre-trained embeddings and contextualization techniques, it can produce coherent and interpretable results without the extensive computational cost required by LLMs, making it more accessible for scenarios with limited resources.

\textbf{Contrastive Learning.} In the contextualized language model era, contrastive learning has been facilitated by sentence embedding models used for topic modelling \cite{oord2018representation,gao2021simcse,reimers2019sentence}. Our method is largely facilitated by properties brought by contrastive learning, including the different self-similarity patterns for functional tokens and semantic tokens; and the usage of contextualized embeddings to serve as candidates for centroid words \cite{ethayarajh2019contextual,xiao2023isotropy}. This is largely inspired by the \textit{``entourage effect''} found in \cite{xiao2023isotropy,xiao2023length}, highlighting that words in the same sentence all follow semantic tokens, wherever they travel, motivating the usage of cluster-specific candidate contextualized embeddings in our work. To the best of our knowledge, previous work in topic modelling has not fully leveraged these useful properties inherent in sentence embeddings brought by contrastive learning.


\section{Conclusion}

We propose an alternative topic modelling approach that 
extends Top2Vec and BERTopic by using two modules: 1) word embeddings contextualized on the dataset and 2) self-similarity scores to filter out functional words and construct meaningful candidate topic words. 

These two modules are straightforward and flexible, making them adaptable components for other topic models. Furthermore, they are independent of dimension reduction and document clustering which enhances CAST's robustness against parameter changes in UMAP and HDBSCAN algorithms. Moreover, these modules could be further enhanced by integrating new embedding models and advanced clustering techniques, thus offering a scalable framework for future developments in topic modelling.

We evaluated CAST on two benchmark datasets and one Twitter dataset, demonstrating superior performance compared to the strong baseline models in identifying coherent and diverse topics.

\section{Limitations and Future Work}

While our proposed approach demonstrates superior performance in generating coherent and diverse topics, as well as handling noisy data, there are a few limitations to consider. First, CAST relies on pre-trained embedding models to capture contextual information. These models may not fully represent all possible linguistic nuances and could introduce biases based on their training data. Furthermore, the computational complexity of CAST may be higher compared to traditional topic modelling techniques, particularly for large datasets, due to the additional steps involved in contextualized word embedding generation (see Table \ref{tab: running time}). Despite these limitations, CAST offers a robust and interpretable approach to topic modelling, leveraging the latest advancements in NLP and neural embeddings.

For future work, one potential direction is to develop contextualized word embeddings that dynamically change based on documents from the clusters instead of the entire corpus. This approach could help tailor word embeddings to specific topics and better capture the nuanced meanings of words in different contexts. In our experiments, we explored this approach and observed that some less meaningful verbs exhibited higher cosine similarities with the topic vector of the cluster. We attribute this phenomenon to two factors: (1) the clusters were too small, which limited the pool of meaningful candidate topic words, and (2) Some words that could represent the topic were not present in the clusters. A more systematic investigation into optimal methods for selecting candidate topic words within clusters is needed.

Furthermore, our findings align with previous criticisms that word co-occurrence-based metrics, such as NPMI, are not well-suited for recently developed neural topic models (see Section \ref{sec: noisy data}). This underscores the need for more appropriate automatic evaluation metrics for neural topic modelling.

\bibliography{custom}

\begin{thebibliography}{35}
\providecommand{\natexlab}[1]{#1}

\bibitem[{Angelov(2020)}]{angelov2020top2vec}
Dimo Angelov. 2020.
\newblock Top2vec: Distributed representations of topics.
\newblock \emph{arXiv preprint arXiv:2008.09470}.

\bibitem[{Barawi et~al.(2017)Barawi, Lin, and Siddharthan}]{barawi2017automatically}
Mohamad~Hardyman Barawi, Chenghua Lin, and Advaith Siddharthan. 2017.
\newblock Automatically labelling sentiment-bearing topics with descriptive sentence labels.
\newblock In \emph{The 22nd International Conference on Applications of Natural Language to Information Systems (NLDB)}, pages 299--312.

\bibitem[{Bellman(1961)}]{Bellman+1961}
Richard~E. Bellman. 1961.
\newblock \href {https://doi.org/doi:10.1515/9781400874668} {\emph{Adaptive Control Processes}}.
\newblock Princeton University Press, Princeton.

\bibitem[{Beyer et~al.(1999)Beyer, Goldstein, Ramakrishnan, and Shaft}]{beyer1999nearest}
Kevin Beyer, Jonathan Goldstein, Raghu Ramakrishnan, and Uri Shaft. 1999.
\newblock When is “nearest neighbor” meaningful?
\newblock In \emph{Database Theory—ICDT’99: 7th International Conference Jerusalem, Israel, January 10--12, 1999 Proceedings 7}, pages 217--235. Springer.

\bibitem[{Blei et~al.(2003)Blei, Ng, and Jordan}]{blei2003latent}
David~M Blei, Andrew~Y Ng, and Michael~I Jordan. 2003.
\newblock Latent dirichlet allocation.
\newblock \emph{Journal of machine Learning research}, 3(Jan):993--1022.

\bibitem[{Bouma(2009)}]{bouma2009normalized}
Gerlof Bouma. 2009.
\newblock Normalized (pointwise) mutual information in collocation extraction.
\newblock \emph{Proceedings of GSCL}, 30:31--40.

\bibitem[{Campello et~al.(2013)Campello, Moulavi, and Sander}]{campello2013density}
Ricardo~JGB Campello, Davoud Moulavi, and J{\"o}rg Sander. 2013.
\newblock Density-based clustering based on hierarchical density estimates.
\newblock In \emph{Pacific-Asia conference on knowledge discovery and data mining}, pages 160--172. Springer.

\bibitem[{Devlin et~al.(2018)Devlin, Chang, Lee, and Toutanova}]{devlin2018bert}
Jacob Devlin, Ming-Wei Chang, Kenton Lee, and Kristina Toutanova. 2018.
\newblock Bert: Pre-training of deep bidirectional transformers for language understanding.
\newblock \emph{arXiv preprint arXiv:1810.04805}.

\bibitem[{Dieng et~al.(2020)Dieng, Ruiz, and Blei}]{dieng2020topic}
Adji~B Dieng, Francisco~JR Ruiz, and David~M Blei. 2020.
\newblock Topic modeling in embedding spaces.
\newblock \emph{Transactions of the Association for Computational Linguistics}, 8:439--453.

\bibitem[{Doogan and Buntine(2021)}]{doogan2021topic}
Caitlin Doogan and Wray Buntine. 2021.
\newblock Topic model or topic twaddle? re-evaluating demantic interpretability measures.
\newblock In \emph{North American Association for Computational Linguistics 2021}, pages 3824--3848. Association for Computational Linguistics (ACL).

\bibitem[{Egger and Yu(2022)}]{egger2022topic}
Roman Egger and Joanne Yu. 2022.
\newblock A topic modeling comparison between lda, nmf, top2vec, and bertopic to demystify twitter posts.
\newblock \emph{Frontiers in sociology}, 7:886498.

\bibitem[{Ethayarajh(2019)}]{ethayarajh2019contextual}
Kawin Ethayarajh. 2019.
\newblock How contextual are contextualized word representations? comparing the geometry of bert, elmo, and gpt-2 embeddings.
\newblock \emph{arXiv preprint arXiv:1909.00512}.

\bibitem[{Gao et~al.(2021)Gao, Yao, and Chen}]{gao2021simcse}
Tianyu Gao, Xingcheng Yao, and Danqi Chen. 2021.
\newblock Simcse: Simple contrastive learning of sentence embeddings.
\newblock \emph{arXiv preprint arXiv:2104.08821}.

\bibitem[{Ghasiya and Okamura(2021)}]{ghasiya2021investigating}
Piyush Ghasiya and Koji Okamura. 2021.
\newblock Investigating covid-19 news across four nations: A topic modeling and sentiment analysis approach.
\newblock \emph{Ieee Access}, 9:36645--36656.

\bibitem[{Grootendorst(2022)}]{grootendorst2022bertopic}
Maarten Grootendorst. 2022.
\newblock Bertopic: Neural topic modeling with a class-based tf-idf procedure.
\newblock \emph{arXiv preprint arXiv:2203.05794}.

\bibitem[{Hoyle et~al.(2021)Hoyle, Goel, Hian-Cheong, Peskov, Boyd-Graber, and Resnik}]{hoyle2021automated}
Alexander Hoyle, Pranav Goel, Andrew Hian-Cheong, Denis Peskov, Jordan Boyd-Graber, and Philip Resnik. 2021.
\newblock Is automated topic model evaluation broken? the incoherence of coherence.
\newblock \emph{Advances in neural information processing systems}, 34:2018--2033.

\bibitem[{Jeon et~al.(2023)Jeon, Yoon, and Sohn}]{jeon2023exploring}
Eunji Jeon, Naeun Yoon, and So~Young Sohn. 2023.
\newblock Exploring new digital therapeutics technologies for psychiatric disorders using bertopic and patentsberta.
\newblock \emph{Technological Forecasting and Social Change}, 186:122130.

\bibitem[{Lau et~al.(2014)Lau, Newman, and Baldwin}]{lau-etal-2014-machine}
Jey~Han Lau, David Newman, and Timothy Baldwin. 2014.
\newblock \href {https://doi.org/10.3115/v1/E14-1056} {Machine reading tea leaves: Automatically evaluating topic coherence and topic model quality}.
\newblock In \emph{Proceedings of the 14th Conference of the {E}uropean Chapter of the Association for Computational Linguistics}, pages 530--539, Gothenburg, Sweden. Association for Computational Linguistics.

\bibitem[{Lin et~al.(2012)Lin, He, Pedrinaci, and Domingue}]{lin2012feature}
Chenghua Lin, Yulan He, Carlos Pedrinaci, and John Domingue. 2012.
\newblock Feature {LDA}: a supervised topic model for automatic detection of web api documentations from the web.
\newblock In \emph{The 11th International Semantic Web Conference}, pages 328--343.

\bibitem[{McInnes et~al.(2018)McInnes, Healy, and Melville}]{mcinnes2018umap}
Leland McInnes, John Healy, and James Melville. 2018.
\newblock Umap: Uniform manifold approximation and projection for dimension reduction.
\newblock \emph{arXiv preprint arXiv:1802.03426}.

\bibitem[{Meng et~al.(2022)Meng, Zhang, Huang, Zhang, and Han}]{meng2022topic}
Yu~Meng, Yunyi Zhang, Jiaxin Huang, Yu~Zhang, and Jiawei Han. 2022.
\newblock Topic discovery via latent space clustering of pretrained language model representations.
\newblock In \emph{Proceedings of the ACM Web Conference 2022}, pages 3143--3152.

\bibitem[{Mikolov et~al.(2013{\natexlab{a}})Mikolov, Chen, Corrado, and Dean}]{mikolov2013efficient}
Tomas Mikolov, Kai Chen, Greg Corrado, and Jeffrey Dean. 2013{\natexlab{a}}.
\newblock Efficient estimation of word representations in vector space.
\newblock \emph{arXiv preprint arXiv:1301.3781}.

\bibitem[{Mikolov et~al.(2013{\natexlab{b}})Mikolov, Sutskever, Chen, Corrado, and Dean}]{mikolov2013distributed}
Tomas Mikolov, Ilya Sutskever, Kai Chen, Greg~S Corrado, and Jeff Dean. 2013{\natexlab{b}}.
\newblock Distributed representations of words and phrases and their compositionality.
\newblock \emph{Advances in neural information processing systems}, 26.

\bibitem[{Mu et~al.(2024)Mu, Dong, Bontcheva, and Song}]{mu2024large}
Yida Mu, Chun Dong, Kalina Bontcheva, and Xingyi Song. 2024.
\newblock Large language models offer an alternative to the traditional approach of topic modelling.
\newblock \emph{arXiv preprint arXiv:2403.16248}.

\bibitem[{Newman et~al.(2010)Newman, Lau, Grieser, and Baldwin}]{newman2010automatic}
David Newman, Jey~Han Lau, Karl Grieser, and Timothy Baldwin. 2010.
\newblock Automatic evaluation of topic coherence.
\newblock In \emph{Human language technologies: The 2010 annual conference of the North American chapter of the association for computational linguistics}, pages 100--108.

\bibitem[{Oord et~al.(2018)Oord, Li, and Vinyals}]{oord2018representation}
Aaron van~den Oord, Yazhe Li, and Oriol Vinyals. 2018.
\newblock Representation learning with contrastive predictive coding.
\newblock \emph{arXiv preprint arXiv:1807.03748}.

\bibitem[{Rahimi et~al.(2024)Rahimi, Mimno, Hoover, Naacke, Constantin, and Amann}]{rahimi-etal-2024-contextualized}
Hamed Rahimi, David Mimno, Jacob Hoover, Hubert Naacke, Camelia Constantin, and Bernd Amann. 2024.
\newblock \href {https://aclanthology.org/2024.findings-eacl.123} {Contextualized topic coherence metrics}.
\newblock In \emph{Findings of the Association for Computational Linguistics: EACL 2024}, pages 1760--1773, St. Julian{'}s, Malta. Association for Computational Linguistics.

\bibitem[{Reimers and Gurevych(2019)}]{reimers2019sentence}
Nils Reimers and Iryna Gurevych. 2019.
\newblock Sentence-bert: Sentence embeddings using siamese bert-networks.
\newblock \emph{arXiv preprint arXiv:1908.10084}.

\bibitem[{Sia et~al.(2020)Sia, Dalmia, and Mielke}]{sia2020tired}
Suzanna Sia, Ayush Dalmia, and Sabrina~J Mielke. 2020.
\newblock Tired of topic models? clusters of pretrained word embeddings make for fast and good topics too!
\newblock \emph{arXiv preprint arXiv:2004.14914}.

\bibitem[{Terragni et~al.(2021)Terragni, Fersini, Galuzzi, Tropeano, and Candelieri}]{terragni2021octis}
Silvia Terragni, Elisabetta Fersini, Bruno~Giovanni Galuzzi, Pietro Tropeano, and Antonio Candelieri. 2021.
\newblock Octis: Comparing and optimizing topic models is simple!
\newblock In \emph{Proceedings of the 16th Conference of the European Chapter of the Association for Computational Linguistics: System Demonstrations}, pages 263--270.

\bibitem[{Wang et~al.(2023{\natexlab{a}})Wang, Prakash, Hoang, Hee, Naseem, and Lee}]{wang2023prompting}
Han Wang, Nirmalendu Prakash, Nguyen~Khoi Hoang, Ming~Shan Hee, Usman Naseem, and Roy Ka-Wei Lee. 2023{\natexlab{a}}.
\newblock Prompting large language models for topic modeling.
\newblock In \emph{2023 IEEE International Conference on Big Data (BigData)}, pages 1236--1241. IEEE.

\bibitem[{Wang et~al.(2023{\natexlab{b}})Wang, Chen, Chen, and Chen}]{wang2023identifying}
Zhongyi Wang, Jing Chen, Jiangping Chen, and Haihua Chen. 2023{\natexlab{b}}.
\newblock Identifying interdisciplinary topics and their evolution based on bertopic.
\newblock \emph{Scientometrics}, pages 1--26.

\bibitem[{Xiao et~al.(2023{\natexlab{a}})Xiao, Li, Hudson, Lin, and Al~Moubayed}]{xiao2023length}
Chenghao Xiao, Yizhi Li, G~Hudson, Chenghua Lin, and Noura Al~Moubayed. 2023{\natexlab{a}}.
\newblock Length is a curse and a blessing for document-level semantics.
\newblock In \emph{Proceedings of the 2023 Conference on Empirical Methods in Natural Language Processing}, pages 1385--1396.

\bibitem[{Xiao et~al.(2023{\natexlab{b}})Xiao, Long, and Al~Moubayed}]{xiao2023isotropy}
Chenghao Xiao, Yang Long, and Noura Al~Moubayed. 2023{\natexlab{b}}.
\newblock On isotropy, contextualization and learning dynamics of contrastive-based sentence representation learning.
\newblock In \emph{Findings of the Association for Computational Linguistics: ACL 2023}, pages 12266--12283.

\bibitem[{Yuan and Eldardiry(2021)}]{yuan-eldardiry-2021-unsupervised}
Chenhan Yuan and Hoda Eldardiry. 2021.
\newblock \href {https://doi.org/10.18653/v1/2021.emnlp-main.147} {Unsupervised relation extraction: A variational autoencoder approach}.
\newblock In \emph{Proceedings of the 2021 Conference on Empirical Methods in Natural Language Processing}, pages 1929--1938, Online and Punta Cana, Dominican Republic. Association for Computational Linguistics.

\end{thebibliography}

\appendix
\section{Running Time}
\label{appen: running time}
In this section, we report the computational cost for each model in Table \ref{tab: running time}. The results were averaged across five independent runs for three different topic counts, totalling 15 runs per model on an NVIDIA A100 GPU. 

The traditional LDA model is faster than the rest neural topic models. TopClus exhibits the highest computational demand. Our proposed approach generally requires more time than BERTopic and Top2Vec due to the additional generation of contextualized word embeddings when running the first time (e.g. around 900$s$ on 20NewsGroup using CAST-MPNET). However, our method's sentence and word embeddings can be saved and reused for future runs, thus reducing the computational time to around 10-20$s$.

\begin{table}[!ht]
\centering
\resizebox{\columnwidth}{!}{
\begin{tabular}{cccc}
\toprule
\textbf{Model} & \textbf{20NewsGroup} & \textbf{BBC News} & \textbf{Elonmusk} \\
\midrule
LDA & $12.52$ & $5.78$ & $4.37$ \\
TopClus & $1675.97$ & $255.3$ & $229.2$ \\
BERTopic-\textit{MiniLM} & $17.48$ & $9.54$ & $13.86$\\
BERTopic-\textit{MPNET} & $16.54$ & $11.3$ & $14.92$\\
Top2Vec-\textit{MiniLM}& $47.82$ & $12.75$ & $25.51$\\
Top2Vec-\textit{MPNET}& $135.3$ & $32.5$ & $48.12$\\
CAST-\textit{MiniLM} & $91.37$ & $21.97$ & $59.12$\\
CAST-\textit{MPNET} & $83.63$ & $20.49$ & $27.74$\\
\bottomrule
\end{tabular}}
\caption{Computation time ($seconds$) for each model across different datasets.}
\label{tab: running time}
\end{table}

\noindent
\setlength{\tabcolsep}{3pt}
\begin{table}[!htbp]
\centering
\resizebox{\columnwidth}{!}{
\begin{tabular}{cccccc}
\hline
\textbf{Datasets} & \textbf{\#Docs} &  \textbf{\#Labels} & \textbf{MCS} & \textbf{\#Topics}\\ \hline
20NewsGroup & $16,309$   &  $20$    & $15$  & $\{10, 20, 30\}$ \\
BBC News   & $2,225$     &   $5$    & $5$   & $\{5, 10, 15\}$  \\
Elon Musk   & $13,998$    &   $-$    & $15$  & $\{10, 20, 30\}$ \\ \hline
\end{tabular}
}
\caption{Data statistics. The \texttt{Min\_Cluster\_Size} (MCS) and \texttt{\#Topics} were determined based on the dataset size, with larger datasets having larger values. We did not report the \texttt{\#Labels} of Elon Musk's Tweets as this dataset has not been human-labelled.}
\label{data-statistics}
\vspace{-3mm}
\end{table}

\section{LLM-Based Metrics Design and GPT-4 Validation}
\label{appen: GPT-4 validation}

\begin{figure*}[!htbp]
    \begin{dialogbox}
    \small
\textbf{Coherence Prompt:}
I will provide you with sets of clusters, where each cluster is described by a list of keywords: [topic\_words]. For each set, evaluate the topic coherence of the clusters based on how interpretable and meaningful the keyword lists are for representing and retrieving distinct topics or subjects. Use this 4-point rating scale:\vspace{0.05in}

\textbf{4 = Keywords are highly coherent and clearly represent a specific, well-defined topic} \\ 
\textbf{3 = Keywords are reasonably coherent and suggest a relatively distinct topic} \\   
\textbf{2 = Keywords lack coherence and make the topic difficult to interpret} \\ 
\textbf{1 = Keywords are essentially random and meaningless for defining any topic} \\
For each cluster set, provide: The cluster number and your rating score, along with a 1-2 sentence explanation, in this format:  

\textbf{Cluster [X]: [score] - [brief explanation]} \\
Then calculate and provide the average cluster score as the "Topic Coherence Score" for that set. \\

\textbf{Diversity Prompt:} I will provide you with sets of clusters, where each cluster is described by a list of keywords: [topic\_words]. Evaluate the diversity of topics represented across all the clusters in each set on a 4-point rating scale: \vspace{0.05in}

\textbf{4 = Extremely diverse topics (clusters cover a very wide range of completely distinct and unrelated topics)} \\
\textbf{3 = High topic diversity (clusters cover many distinct topics with little overlap)} \\  
\textbf{2 = Moderate topic diversity (clusters cover some distinct topics but also significant overlap)} \\ 
\textbf{1 = Low topic diversity (clusters cover highly repetitive clusters with very little distinction in topics covered)} \\ 
For each set, provide your score along with a brief explanation justifying the rating. The response should be in the following format: 
\textbf{Set [name]: [score] [Explanation]}. \\

    \end{dialogbox}
    \caption{The LLM-based prompts to evaluate the coherence and diversity of the models.}
    \label{fig:prompt}
\end{figure*}

\begin{table*}[h]
\centering
\scalebox{0.9}{
\begin{tabular}{cccccccccccccccccc}
\toprule
\multirow{2}{*}{\textbf{TC}} & \textbf{GPT-4} & $3.33$ & $3.98$ & $3.57$ & $3.83$ & $2.52$ & $3.32$ & $2.73$ & $3.63$ & $3.8$ & $3$ & $2.5$ & $3.3$ & $3.93$ & $2.7$ & $3.55$ & $3.5$ \\
& \textbf{Human} & $3$    & $3.85$ & $3.6$ & $3.5$ & $2.15$ & $2.7$ & $2.4$ & $3.45$ & $3.8$  & $1.8$  & $1.6$ & $1.85$ & $3.9$ & $2.5$  & $3.4$ & $3.2$ \\
\bottomrule
\toprule
\multirow{2}{*}{\textbf{TD}} & \textbf{GPT-4} & $2.33$ & $3.33$ & $3$   & $4$   & $2.67$ & $3$   & $3$   & $2.33$ & $3.33$ & $2.33$ & $1$   & $1$    & $4$   & $2.33$ & $2.3$ & $3$   \\
& \textbf{Human}& $2$    & $3$    & $4$   & $3.5$ & $2$  & $2$ & $2$ & $2$  & $3$  & $1$    & $1.5$ & $1$    & $3.5$ & $2$    & $3$   & $2$   \\ 
\bottomrule
\end{tabular}}
\caption{Evaluation results by GPT-4 and human for Topic Coherence (TC) and Topic Diversity (TD) on a subset data of 20NewsGroups.}
\label{tab: human evaluation}
\end{table*}

In this section, we present more detailed information about the design of LLM-based metrics and the validation for the use of GPT-4. 

We design the prompts to evaluate the models using 4-point rating scales as shown in Figure \ref{fig:prompt}. The coherence prompt assesses how well the topic words can represent and retrieve documents about a particular topic, where 4 indicates an extremely coherent topic and 1 indicates a meaningless topic. The diversity prompt assesses how diverse the identified topics are, with 4 indicating extremely diverse topics and 1 indicating highly overlapping topics. 

To validate our GPT-4-based automatic evaluation approach, we conducted a parallel human evaluation. We recruited two linguistics experts and provided them with instructions and examples of high and low-quality outcomes. Before formal evaluation, we conducted a calibration exercise to ensure consistent evaluation standards. The two experts then independently assessed randomly selected results from each model's run on the 20NewsGroups dataset. Their scores were averaged and then compared with GPT-4 evaluations (see Table \ref{tab: human evaluation}). We used the default ChatGPT web chat (the \texttt{gpt-4-turbo} version) in our experiment. The Pearson correlation coefficients between GPT-4 and human evaluations were 0.869 for TC and 0.732 for TD, supporting the efficacy of GPT-4 for automatic evaluation. 

As the prompt design is beyond the focus of our paper, we only used zero-short prompting. In the future, researchers could use more advanced prompting to achieve a better correlation with human evaluation.

\section{Implementation Details}
\label{appen: implementation}

In our experiment, we used UMAP to reduce the dimensionality of the document embeddings and HDBSCAN to cluster the reduced embeddings as suggested in Top2Vec and BERTopic. UMAP has the ability to preserve both local and global structure and be scalable to large datasets \citep{mcinnes2018umap, angelov2020top2vec, grootendorst2022bertopic}. HDBSCAN adopts a minimum spanning tree algorithm based on data point densities to identify clusters and uses a soft-clustering method to identify noise and outlier points.

To ensure a fair comparison, we kept and fixed the UMAP and HDBSCAN parameters across the models. For UMAP, \texttt{n\_neighbors} determines how UMAP balances local and global structure by controlling the size of the local neighbourhood considered when learning the manifold, with lower values focusing on local details and higher values capturing broader data patterns. We set it to the default 15 in our experiment. We used the default \texttt{metric = cosine} to calculate the distance in the ambient space of the input data. The \texttt{n\_components} controls the reduced dimensionality. We set it to 5 as Top2Vec suggests this value has the best results for the downstream task of density-based clustering. 

For HDBSCAN, \texttt{min\_cluster\_size} represents the smallest size that should be considered a cluster by the algorithm. We adjusted the \texttt{min\_cluster\_size} based on the dataset size (see Table \ref{data-statistics}), with larger datasets having a larger \texttt{min\_cluster\_size} value.

\section{Topic Words Example}
\label{appen: detailed_twords}
In this section, we choose 20NewsGroups as an example to show detailed topic words generated by each model with \texttt{\#Topic=10}. The generated topic words are presented to illustrate the effectiveness of each model in capturing coherent and diverse topics within the dataset.

\noindent
\begin{table*}[!ht]
\centering
\scalebox{0.68}{
\begin{tabular}{ll}
\toprule
\textbf{Models} &{\textbf{20NewsGroups}} \\
\midrule
\multirow{10}{*}{\makecell{LDA}} 
&  wire, ground, circuit, problem, work, power, cable, car, box, hot \\
&  image, bit, file, card, color, display, driver, work, window, monitor\\
&  people, book, time, make, israeli, give, kill, jewish, find, quote \\
&  key, law, government, public, gun, protect, case, chip, state, number \\
&  good, make, time, win, car, bike, thing, give, back, lose \\
&  file, system, run, program, window, include, software, application, list, drive \\
&  year, people, time, space, report, study, day, make, woman, child \\
&  game, year, team, player, good, play, make, time, win, season \\
&  people, make, thing, claim, religion, argument, word, church, question, atheist \\
&  make, president, people, time, work, job, thing, decision, question, year\\

\midrule
\multirow{10}{*}{\makecell{TopClus}} 
&  food, music, land, water, range, fuel, height, temperature, rock, gas \\
&  make, establish, bring, give, begin, organize, conclude, prepare, accept, feel\\ 
&  government, gospel, christianity, church, catholic, worship, civilian, sex, religious, morality \\
&  shoot, hit, move, play, put, live, set, push, playoff, jump \\
&  maximum, typical, good, long, future, large, normal, half, big, small \\
&  people, kid, person, year, season, boy, time, man, day, father \\
&  dat, del, black, graphic, internal, open, red, automatic, automatically, color \\
&  method, study, permission, conclusion, plan, decision, feeling, thinking, reason, result \\
&  newspaper, verse, speech, number, writing, read, letter, language, reading, post \\
&  box, system, hardware, card, controller, device, computer, keyboard, connector, customer \\

\midrule
\multirow{10}{*}{\makecell{BERTopic-\textit{MiniLM}}} 
& people, make, file, time, government, key, give, find, image, thing\\
& car, space, bike, good, make, launch, time, year, system, price\\
& game, team, play, player, year, season, win, score, good, hit\\
& drive, scsi, card, modem, monitor, driver, color, disk, problem, work\\
& window, file, mouse, run, motif, program, application, problem, work, manager\\
& font, printer, print, laser, window, problem, driver, paper, quality, page\\
& point, plane, line, edge, surface, face, algorithm, normal, circle, center\\
& driver, sound, file, play, window, work, instal, card, set, speaker\\
& sort, number, utility, worth, thing, suggestion, understand, sound, label, doesn\\
& insurance, company, fault, accident, car, lawyer, pay, stay, drop, damage\\

\midrule
\multirow{10}{*}{\makecell{BERTopic-\textit{MPNET}}} 
& people, make, year, time, game, good, give, team, thing, point\\
& file, key, drive, system, image, chip, program, disk, bit, card\\
& car, bike, engine, ride, mile, drive, battery, good, buy, road\\
& window, file, manager, program, run, error, problem, include, compile, server\\
& sale, offer, sell, condition, price, interested, good, pay, include, tape\\
& mail, email, address, list, send, mailing, post, advance, reply, response\\
& font, printer, print, laser, window, character, driver, problem, page, paper\\
& color, plastic, blue, draw, bit, paint, board, green, line, display\\
& mouse, keyboard, driver, problem, work, window, button, load, serial, ball\\
& motif, widget, application, run, server, code, problem, region, window, call\\

\midrule
\multirow{10}{*}{\makecell{Top2Vec-\textit{MiniLM}}} 
& hardware, motherboard, workstation, ram, processor, configuration, modem, mhz, driver, computer\\
& behavior, act, consequence, incident, evidence, investigation, law, conclusion, statement, court\\
& government, enforcement, civilian, security, protect, law, policy, military, police, survey\\
& belief, christianity, religion, religious, church, atheism, doctrine, faith, philosophy, worship\\
& program, software, file, application, screen, programming, window, workstation, compile, code\\
& baseball, league, hockey, playoff, pitch, sport, game, ball, team, bat\\
& survey, demand, design, technology, requirement, activity, proposal, development, interest, suggestion\\
& vehicle, car, truck, motorcycle, motor, engine, bike, dealer, road, drive\\
& sale, purchase, buy, sell, price, shipping, vendor, cheap, dealer, market\\
& mail, mailing, email, contact, send, address, message, telephone, exchange, inform\\

\bottomrule
\text{Continued on next page} & {}
\end{tabular}
}
\end{table*}

\begin{table*}[ht]
\centering
\scalebox{0.7}{
\begin{tabular}{ll}

\toprule
\multirow{10}{*}{\makecell{Top2Vec-\textit{MPNET}}} 
& government, political, crime, inform, topic, police, administration, implement, document, organization\\
& inform, organization, communication, implement, management, introduction, topic, medicine, business, document\\
& religion, christianity, philosophy, religious, scripture, belief, literature, society, doctrine, culture\\
& implement, program, software, documentation, implementation, format, programming, application, document, technology\\
& hardware, motherboard, computer, monitor, processor, vga, cpu, ram, workstation, device\\
& vehicle, car, engine, truck, auto, motorcycle, motor, manual, bike, drive\\
& technology, device, tech, implement, interface, computer, implementation, software, hardware, modem\\
& baseball, hockey, pitch, league, sport, game, win, playoff, lose, team\\
& program, compile, window, ide, issue, configuration, problem, application, instal, software\\
& sale, sell, purchase, price, buy, market, vendor, trade, mailing, exchange\\

\midrule
\multirow{10}{*}{\makecell{CAST-\textit{MiniLM}}} 
& game, team, baseball, player, hockey, sport, score, league, ball, season\\
& car, vehicle, truck, motor, auto, road, motorcycle, engine, bike, dealer\\
& belief, religion, faith, truth, god, argue, religious, christianity, holy, church\\
& medicine, health, medical, practice, encourage, patient, prove, doctor, disease, concern\\
& encrypt, encryption, secure, security, privacy, protect, prove, deny, scheme, secret\\
& mission, orbit, space, project, launch, proposal, technical, technology, satellite, engineering\\
& mail, email, reply, message, letter, respond, post, response, request, announcement\\
& gun, firearm, criminal, weapon, enforcement, crime, police, violent, shoot, armed\\
& israeli, arab, peace, jewish, terrorist, deny, oppose, civilian, argue, prove\\
& drive, disk, scsi, ide, motherboard, hardware, device, tech, rom, controller\\

\midrule
\multirow{10}{*}{\makecell{CAST-\textit{MPNET}}} 
& game, score, team, play, player, season, league, shot, goal, tie\\
& car, rear, vehicle, brake, engine, auto, gear, ride, motor, wheel\\
& secure, security, scheme, encrypt, key, ensure, secret, protect, privacy, private\\
& patient, medical, treat, medicine, treatment, doctor, disease, health, hospital, sick\\
& disk, scsi, ide, controller, device, bus, hardware, rom, tech, pin\\
& launch, mission, space, flight, orbit, rocket, satellite, solar, moon, facility\\
& doctrine, faith, church, holy, gospel, christian, pray, religion, scripture, god\\
& mail, email, mailing, send, message, post, newsgroup, request, posting, inform\\
& sale, sell, price, shipping, warranty, trade, dealer, product, firm, market\\
& audio, signal, noise, circuit, input, channel, equipment, listen, design, voice\\

\midrule
\multirow{10}{*}{\makecell{CAST\textit{MiniLM}$*$}} 

& game, team, play, baseball, player, hockey, sport, score, league, ball \\
& suggest, give, related, treat, medicine, involve, situation, explain, helpful, provide \\
& encrypt, provide, encryption, implement, secure, establish, possibly, security, ensure, introduce\\
& orbit, provide, flight, mission, space, achieve, suggest, launch, implement, accomplish\\
& bike, motorcycle, ride, vehicle, wheel, SEP, road, truck, rear, car\\
& car, vehicle, auto, bike, buy, engine, excellent, dealer, motor, truck\\
& drive, disk, scsi, ide, motherboard, hardware, device, install, tech, rom\\
& printer, print, font, disk, paper, display, make, driver, fix, software\\
& turkish, armenian, massacre, soviet, genocide, russian, seek, muslim, establish, attempt\\
& homosexual, gay, sexual, homosexuality, sex, discuss, argue, suggest, behavior, issue\\

\midrule
\multirow{10}{*}{\makecell{CAST-\textit{MPNET}$*$}}
& game, score, team, play, player, hit, win, season, league, shot \\
& secure, security, encrypt, scheme, key, secret, ensure, protect, assure, privacy\\
&patient, medical, treat, medicine, treatment, doctor, disease, health, care, affect\\
&car, engine, auto, vehicle, rear, dealer, motor, brake, gear, mile\\
&drive, disk, scsi, ide, controller, hard, instal, device, bus, hardware \\
& doctrine, faith, church, holy, christian, gospel, pray, religion, god, christianity\\
& launch, mission, space, flight, orbit, rocket, satellite, solar, moon, achieve \\
& monitor, vga, card, video, screen, display, driver, compatible, resolution, graphic \\
& occupy, territory, soldier, israeli, peace, war, land, civilian, attack, foreign \\
& bike, ride, motorcycle, rear, gear, wheel, brake, tire, front, mile \\

\bottomrule
\end{tabular}
}
\caption{The top 10 topic words generated by each model on 20NewsGroups with \#Topics=10. $^*$ means the model does not use the self-similarity module to filter words.}
\label{tab:detailed_twords}
\end{table*}
\FloatBarrier

\end{document}